\documentclass{article}
\usepackage{graphicx}
\usepackage{amsmath}
\usepackage{url}
\usepackage{xcolor}
\usepackage{natbib}

\newenvironment{eqnnon}{\begin{equation*}}{\end{equation*}}

\newcommand{\revision}{}

\begin{document}

\title{Solving for multi-class using orthogonal coding matrices}

\author{Peter Mills}

\maketitle

\begin{center}
  \includegraphics[width=0.6\textwidth]{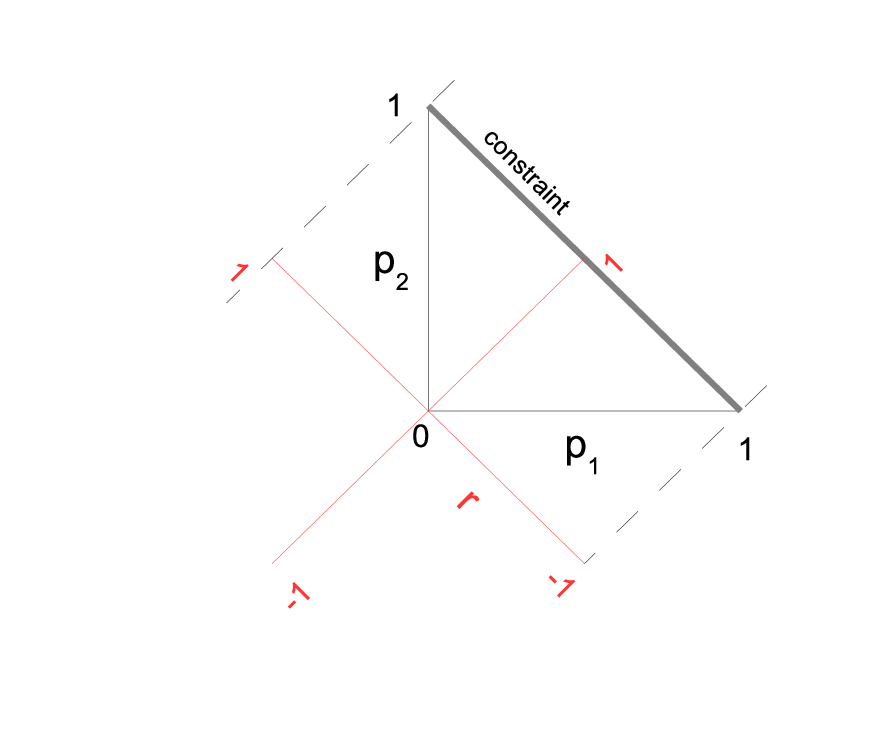}
\end{center}

\section*{Abstract}

	A common method of generalizing binary to multi-class classification
is the error correcting code (ECC).
ECCs may be optimized in a number of ways, for instance by making them
orthogonal.
Here we test two types of orthogonal ECCs on seven different datasets using
three types of binary classifier and compare them with three
other multi-class methods: 1 vs. 1, one-versus-the-rest and random
ECCs.
The first type of orthogonal ECC, in which the codes contain no zeros,
admits a fast and simple method of solving for the probabilities.

Orthogonal ECCs are always more accurate than random ECCs as predicted by
recent literature.
Improvments in uncertainty coefficient (U.C.)
range between 0.4--17.5\% (0.004--0.139, absolute),
while improvements in Brier score between 0.7--10.7\%.
Unfortunately, orthogonal ECCs are rarely more accurate than
1 vs. 1.
Disparities are worst when the methods are paired with logistic regression,
with orthogonal ECCs never beating 1 vs. 1.
When the methods are paired with SVM, the losses are less significant,
peaking at 1.5\%, relative, 0.011 absolute in uncertainty coefficient
and 6.5\% in Brier scores.

Orthogonal ECCs are always the fastest of the five multi-class
methods when paired with linear classifiers.
When paired with a piecewise linear classifier, whose classification
speed does not depend on the number of training samples, classifications using
orthogonal ECCs were always more accurate than the other
methods\footnote{That is: than either 1 vs. rest or a random ECC}
 and also faster than 1 vs. 1. 
Losses against 1 vs. 1 here were higher, peaking at 1.9\%
(0.017, absolute), in U.C. and 39\% in Brier score.
Gains in speed ranged between 1.1\% and over 100\%.
Whether the speed increase is worth the penalty in accuracy
will depend on the application.

\section{Introduction}

Many methods of statistical classication can only discriminate between two classes. 
Examples include lineear classifiers such as perceptrons and logistic regression \citep{Michie_etal1994}, 
piecewise linear classifiers \citep{Herman_Yeung1992,Mills2018},
as well as support vector machines \citep{Mueller_etal2001}.
There are many ways of generalizing binary classification to 
multi-class {\revision and the number of possibilities increases exponentially with
the number of classes.}

{\revision One should distinguish between multi-class methods that use only a subset
of the binary classifiers, adding more as the algorithm narrows down the
class, and those that use all of the binary classifiers, combining the results
or solving for the class probabilities.
In the former category, we have hierarchical multi-class classifiers 
such as decision trees \citep{Cheong_etal2004,Lee_Oh2003} and decision
directed acyclic graphs (DDACs) \citep{Platt_etal2000}.
In the latter category,
two common methods are one-versus-one (1 vs. 1) and one-versus-the-rest (1 vs. rest) \citep{Hsu_Lin2002}.
These in turn generalize to error-correcting codes (ECCs) \citep{Dietterich_Bakiri1995}.}

{\revision Early experiments with ECCs used random codes:
the assumption is that if the codes are long enough (there are enough binary classifiers)
they will adequately span the classes.
Later work focused on optimizing the design of the codes: what type of
codes will best span the classes and produce the most accurate results?
Here we can also distinguish between two types: those that use the data
to help design the codes \citep{Crammer_Singer2002,Zhou_etal2008,Zhong_Cheriet2013}
and those that are independent of the data but use the mathematical 
properties of the codes themselves to aid in their construction \citep{Allwein_etal2000,Windeatt_Ghaderi2002,Zhou_etal2019}.
It is these latter type of optimized error-correcting codes we turn to in this note.}

In error-correcting coding, there is a coding matrix, $A$, that specifies
how the set of multiple classes is partitioned for each binary classifier.
{\revision For a given column, 
if members of the $j$th class are to be labeled $-1/+1$ for the
binary classifier, then the $j$th row is assigned a $-1/+1$.
If the $j$th class is left out, then the $j$th row is assigned a $0$.}
Typically, the class of the test point is determined by the distance between
a row in the matrix and a vector of binary {\it decision functions}:
\begin{equation}
	c(\vec x) = \arg \min_i | \vec a_i - \vec r(\vec x) |
	\label{min_dist}
\end{equation}
where $\vec a_i\in \lbrace -1,0,+1 \rbrace$ 
is the $i$th row of the coding matrix and 
$\vec r$ is a vector of decision functions at {\it test point}, $\vec x$.
{\revision
If we take the upright brackets as a Euclidean distance we can expand (\ref{min_dist}) as follows:
\begin{eqnnon}
	c = \arg \min_i \sum_j \left ( | \vec a_i | + | \vec r | - 2 \vec a_i  \cdot \vec r \right ) 
\end{eqnnon}
Since $| \vec r |$ is constant over $i$, it may be removed from the expression.
Also, for the purposes of this note,
each row of $A$ will be given the same number of non-zero entries, hence:
\begin{eqnnon}
	| \vec a_i | = | \vec a_j | = const.
\end{eqnnon}
This is most evident for the case in which each binary classifier partitions
all of the classes so that there are no zeros in $A$
as is the case for the one-versus-the-rest partitioning.}
Then (\ref{min_dist}) reduces to a {\it voting} solution:
\begin{equation}
	c = \arg \max A \vec r \label{voting}
\end{equation}
Both \citet{Allwein_etal2000} and \citet{Windeatt_Ghaderi2002} show that to
maximize the accuracy of an ECC, the distance
between each row, $|\vec a_i - \vec a_j|_{i \ne j}$, should be maximized.
Using the above assumptions, this reduces to:
\begin{eqnnon}
	\min |\vec a_i \cdot \vec a_j|_{i \ne j}
\end{eqnnon}
{\revision Note the absolute value prevents degenerate rows.}
In other words, the coding matrix, $A$, should be orthogonal.

In this note, we describe a fast and simple algorithm that uses orthogonal ECCs to solve for the conditional probabilites in multi-class
classification.
{\revision
There are three reasons to require the conditional probabilities:
\begin{enumerate}
	\item Probabilities provide useful extra information, specifically how accurate a given classification is, in absence of knowledge of its true value.
	\item The relationship between the binary probabilities and the
		multi-class probabilities derives uniquely and rigorously from probability theory.
	\item Binary classifiers that do not return calibrated probability estimates, 
		but nonetheless supply a continuous decision function,
		are easy to recalibrate so that the decision function more closely resembles a probability \citep{Jolliffe_Stephenson2003,Platt1999}.
\end{enumerate}
Two types of orthogonal ECCs along with three other multi-class 
methods--1 vs. 1, 1 vs. the rest, and random ECCs--will be tested on seven
different datasets using three 
different binary classifiers--logistic regression, support vector machines (SVM), and piece-wise linear--to 
see how they compare in terms of classification speed, classification accuracy and accuracy of the conditional probabilities.}

\section{Algorithm}

\label{algorithm}

We wish to design a set of $m$ binary classifiers, each of which return a 
decision function:
\begin{eqnnon}
r_j(\vec x) = P_j(-1 | \vec x) - P_j(+1 | \vec x)
\label{rdef}
\end{eqnnon}
where $P_j(c | \vec x)$ is the conditional probability of the $c$th class of
the $j$th classifier.
Each binary classifier partitions a set of $m$ classes such that for a
given test point, $\vec x$:
\begin{eqnnon}
	\sum_{i=1}^m a_{ij} p_i = r_j; ~~~ j=[1..n]
\label{multiclass}
\end{eqnnon}
where $A=\lbrace a_{ij} \in \lbrace -1, +1 \rbrace  \rbrace$ is a {\it coding
matrix} for which each code partitions all of the classes
and $p_i = p(i | \vec x)$ is the 
conditional  probability of the $i$th class.
In vector notation:
\begin{equation}
	A^T \vec p = \vec r \label{inverse}
\end{equation}
This result derives from the fact that the class probabilities are
additive \citep{Kong_Dietterich1997}.
The more general case where a class can be excluded, that is the coding 
may include zeroes, $a_{ij} \in \lbrace -1, 0, +1\rbrace$,
{\revision will be treated in the next section.}

Note that this assumes that the binary decision functions, $\vec r$,
estimate the conditional probabilities perfectly.
In practice
there are a set of constraints that must be enforced
because $\vec p$ is only allowed to take on certain values.
Thus, we wish to solve the following minimization problem:
\begin{equation}
	\arg \min_{\vec p} | A^T \vec p - \vec r | \label{minimization}
\end{equation}
\begin{eqnarray}
	\sum_{i=1}^m p_i & = & 1 \label{normalization}\\
	p_i & \ge & 0; ~~~ i=[1..m] \label{nonnegative}
\end{eqnarray}

If $A$ is orthogonal,
\begin{eqnnon}
	A A^T = n I
	\label{orthogonal}
\end{eqnnon}
where $I$ is the $m \times m$ identity matrix,
then the unconstrained minimization problem is easy to solve. 
Note that the voting solution in (\ref{voting}) is now equivalent to
the inverse solution in (\ref{inverse}).
This allows us to determine the class easily, but we also wish to solve for
the probabilities, $\vec p$, so that none of the constraints in 
(\ref{normalization}) or (\ref{nonnegative}) are violated.

The orthogonality property allows us to reduce the minimization problem 
in (\ref{minimization}) to something much simpler:
\begin{eqnnon}
	\arg \min_{\vec p} | \vec p - \vec p_0 |
\end{eqnnon}
where $\vec p_0 = A \vec r/n$ with the constraints in (\ref{normalization}) and
(\ref{nonnegative}) remaining the same.
Because the system has been rotated and expanded, the non-negativity 
constraints in (\ref{nonnegative}) remain orthogonal, meaning they are 
independent: enforcing one by setting one of the probabilities to zero, 
$p_k=0$ for example, shouldn't otherwise affect the solution.
This still leaves the normalization constraint in (\ref{normalization}):
the problem, now strictly geometrical, is comprised of finding the point nearest $p_0$ on the diagonal hyper-surface that bisects the unit hyper-cube.

Briefly, we can summarize the algorithm as follows:
1. move to the nearest point that satisfies the normalization constraint,
(\ref{normalization}); 2. if one or more of the probabilities is negative,
move to the nearest point that satisfies both 
the normalization constraint
and the non-negativity constraints, (\ref{nonnegative}), for the negative probabilities;
3. repeat step 2.
More formally, let $\vec 1$ be a vector of all $1$'s:
\begin{itemize}
	\item $i:=0$; $m_0:=m$
	\item while $\exists k \, p_{ik} < 0 \lor \vec p_i \cdot \vec 1 \ne 1$:
	\begin{itemize}
		\item if $\vec p_i \cdot \vec 1 \ne 1$ then 
		$\vec p_{i+1} := \vec p_i + (\vec p_i \cdot \vec 1 - 1)/m_i$
		\item let $K$ be the set of $k$ such that $p_{i+1,k} < 0$
		\item for each $k \in K$:
		\begin{itemize}
			\item $p_k:=0$
			\item Remove $k$ from the problem
		\end{itemize}
		\item $m_{i+1}:=m_i-|K|$
		\item $i:=i+1$
	\end{itemize}
\end{itemize}

Note that resultant direction vectors for each step form an orthogonal set.
For instance, suppose $m_0=4$ and after enforcing the normalization constraint,
the first probability is less than zero, $p_{1,1} < 0$,
then the direction vectors for the two motions are:
\begin{eqnnon}
	\frac{1}{2}[1, 1, 1, 1] \cdot \frac{1}{2\sqrt{3}} [-3, 1, 1, 1] = 0
\end{eqnnon}

More generally, consider the following sequence of vectors:
\begin{eqnnon}
	v_{ij} = \frac{1}{\sqrt{(m-i)^2-2(m-i-1)}} \left \lbrace \begin{array}{rl}
			0; & j < i \\
			-m+i+1; & j=i \\
			1; & j > i
		\end{array} \right .
\end{eqnnon}
where $i \in [1, m]$ and $j \in [1, m]$. \citep{Boyd_Vandenberghe2004}
A nice feature of this method, in addition to being fast,
is that it is divided into two stages: a solution stage and a normalization stage.

\section{Constructing the coding matrix}

\label{construction}

Finding an $A$ such that $A A^T = n I$ and $a_{ij} \in \lbrace -1, 1, \rbrace$
is quite a difficult combinatorial problem.
{\revision When zeros are added in, $a_{ij}\in \lbrace -1, 0, 1\rbrace$,
it becomes even more difficult.}
Work in signal processing may be of limited applicability because coding
matrices are typically comprised of $0$'s and $1$'s 
rather than $-1$'s and $+1$'s \citep{Hedayat_etal1999,Panse_etal2014}.
In our case, a further restriction is that
columns must contain both positive and negative elements, or:
\begin{equation}
	\sum_{i=0}^m a_{ij} \ne \sum_{i=0}^m |a_{ij}|;  ~~~ j=[1..n] \label{restriction}
\end{equation}

A simple method of designing an orthogonal $A$ is using harmonic series.
Consider the following matrix for six classes ($m=6$) 
and eight binary classifiers ($n=8$):
\begin{equation}
	A = \left [ \begin{array}{rrrrrrrr}
			 1 & 1 & 1 & 1 & 1 & 1 & 1 & 1 \\
			-1 & -1 & -1 & -1 & 1 & 1 & 1 & 1 \\
			-1 & -1 & 1 & 1 & -1 & -1 & 1 & 1 \\
			-1 & 1 & -1 & 1 & -1 & 1 & -1 & 1 \\
			 1 &  1 & -1 & -1 & -1 & -1 & 1 & 1 \\
			-1 & 1 & 1 & -1 & -1 & 1 & 1 & -1 
	\end{array} \right ]
	\label{harmonic}
\end{equation}
This will limit the size of $m$ relative to $n$; more precisely:
$m \le \lfloor 2 \log_2 n \rfloor$. Moreover, only certain values of $n$
will be admitted: $n=2^t$ where $t$ is a whole number.

The first three rows in (\ref{harmonic}) comprise a Walsh-Hadamard code \citep{Arora_Barak2009}:
all possible permutations are listed.
A square ($n=m$) orthogonal coding matrix is called a Hadamard matrix
\citep{Sylvester1867}.
It can be shown that besides $n=1$ and $n=2$, only Hadamard matrices of size
$n=4t$ exist,  
and it is still unproven that examples exist for all values of $t$
\citep{Hedayat_Wallis1978}.
A very simple, recursive method exists to generate matrices of size $n=t^2$ 
\citep{Hedayat_Wallis1978} but cannot be made to have the property in (\ref{restriction})
since the matrix includes both a row and column of only ones.
Such a matrix will include a ``harmonic series'' of the same type as in
(\ref{harmonic}).


{\revision Two types of orthogonal coding matrices are tested in this note.
The first type includes no zeros and is generated
using a ``greedy'' algorithm.}
We choose $n$ to be the smallest multiple of $4$ equal to or larger than $m$.
and start with an empty matrix.
Candidate vectors containing both positive and negative elements 
are chosen at random to comprise a row of the matrix but never repeated.
If the candidate vector is orthogonal to existing rows, then it is added to the matrix.
New candidates are tested until the matrix is filled or we run out of permutations.
A full matrix is almost always returned especially if $m<n$.
The matrix is then checked to ensure that 
each column contains both positive and negative elements.
Note that the whole process can be repeated as many times as necessary.
{\revision An eight-class example follows:
\begin{eqnnon}
	A = \left [ \begin{array}{rrrrrrrr}
		1  & -1 & 1 & 1 & -1 & -1 & 1 & -1 \\
		1 & -1 & -1 & 1 & -1 & 1 & -1 & 1 \\
		1 & -1 & -1 & -1 & 1 & 1 & 1 & -1 \\
		1 & 1 & 1 & 1 & 1 & 1 & -1 & -1 \\
		1 & 1 & 1 & -1 & -1 & 1 & 1 & 1 \\
		-1 & -1 & 1 & 1 & 1 & 1 & 1 & 1 \\
		-1 & -1 & 1 & -1 & -1 & 1 & -1 & -1 \\
		1 & -1 & 1 & -1 & 1 & -1 & -1 & 1 
	\end{array} \right ]
\end{eqnnon}
This type of coding matrix can be solved using the algorithm described
in Section \ref{algorithm}, above.}

{\revision
\begin{table}
	\caption{Table showing parameters chosen for the second type of
	orthogonal coding matrix: for the number of classes, $m$,
	the initial length of the code, $n_0$, and the number of non-zero
	values in each code, $|\vec a_i|$ ($i=1..m$), are given. 
	Note: $n_0 \approx m \log_2 m$.}\label{ortho_param}
	\begin{tabular}{l|ll}
		$m$ & $n_0$ & $|\vec a_i|$ \\
		\hline
		4 & 7 & 4 \\
		6 & 12 & 6 \\
		7 & 15 & 7 \\
		8 & 17 & 8 \\
		9 & 20 & 9 \\
		10 & 23 & 10 \\
	\end{tabular}
\end{table}
}

{\revision
The other type of orthogonal coding matrix to be tested in this note
includes zeros.
The construction is similar except now the matrix is allowed to take on values
of zero while the number of non-zero values (-1 or +1) is kept fixed.
A size is chosen for the matrix typically larger than the number of
classes while the resulting matrix will normally be somewhat smaller since
degenerate and fixed value columns (a correctly-trained binary classifier would always return the same value) are removed.
The parameters chosen for each class size are shown in Table
\ref{ortho_param}.}

{\revision
Coding matrices of this type were generated by pure, brute force with no
attempt to track previous trials.
An example coding matrix for six classes is shown below.
Redundant columns have been greyed out.
\begin{eqnnon}
	A = \left [ \begin{array}{rrrrrrrrrrrr}
		-1  &  0 & -1 &  0 &  1 &  0 &  0 &  1 & \color{gray}-1 & \color{gray}0 & \color{gray}0 & \color{gray}-1 \\
		 0  &  1 & -1 &  0 & -1 & -1 &  0 & -1 & \color{gray}-1 & \color{gray}0 & \color{gray}0 &  \color{gray}0 \\
		 -1 &  0 &  1 &  0 &  0 &  0 & -1 & -1 & \color{gray}0 & \color{gray}-1 & \color{gray}0 & \color{gray}-1 \\
		  1 &  0 &  0 &	-1 &  0 & -1 & -1 &  1 & \color{gray}0 & \color{gray}-1 & \color{gray}0 &  \color{gray}0 \\
		  0 & -1 & -1 &  0 & -1 &  1 & -1 &  0 & \color{gray}0 &  \color{gray}0 & \color{gray}1 &  \color{gray}0 \\
		  0 & -1 &  0 &  1 &  0 & -1 &  1 &  0 & \color{gray}0 & \color{gray}-1 & \color{gray}1 &  \color{gray}0 
	\end{array} \right ]
\end{eqnnon}
This type of orthogonal ECC is solved using a general, iterative,
constrained, linear least-squares solver \citep{Lawson_Hanson1995}.}

More work will need to be done to find efficient methods
of generating these matrices
if they are to be applied efficiently to problems with a large number of classes.

\section{Results}

\begin{table*}
{\revision
\caption{Total classification time, solution time, uncertainty coefficient and Brier score for seven different datasets using five different coding matrices: 1 vs. 1, 1 vs. the rest, randoms, orthogonal with no zeros, and orthogonal with zeros. Logistic regression is used as the base binary classifier.}\label{class_results_lin}
\begin{tabular}{ll|llll}
\hline
Dataset & Method & time (s) & sol. only (s) & U.C. & Brier score \\
\hline\hline
	pendigits & 1 vs. 1 & $       0.489\pm   0.006$ & $0.410\pm0.004$ & $     \mathbf{0.956\pm   0.006}$ & $   \mathbf{0.0566\pm   0.003}$\\
 & 1 vs. rest & $       0.118\pm   0.0042$ & $0.0823\pm0.0011$ & $       0.864\pm    0.008$ & $    0.113\pm   0.002$\\
 & ECC & $        0.18\pm    0.01$ & $0.136\pm0.007$ & $     0.723\pm    0.026$ & $    0.180\pm   0.008$\\
	& Ortho. 1 & $       \mathbf{0.048\pm   0.004}$ & $\mathbf{0.01095\pm8e-5}$ & $     0.785\pm   0.010$ & $    0.172\pm   0.002$\\
 & Ortho. 2 & $       0.24\pm    0.01$ & $0.185\pm0.010$ & $     0.862\pm     0.010$ & $    0.123\pm    0.009$\\
\hline
	sat & 1 vs. 1 & $       0.092\pm   0.004$ & $0.067\pm0.001$ & $     \mathbf{0.736\pm   0.009}$ & $    \mathbf{0.176\pm   0.004}$\\
 & 1 vs. rest & $       0.033\pm   0.0048$ & $0.0202\pm2e-4$ & $     0.677\pm   0.007$ & $    0.204\pm   0.002$\\
 & ECC & $       0.043\pm   0.0048$ & $0.0274\pm6e-4$ & $     0.637\pm    0.025$ & $    0.217\pm   0.009$\\
	& Ortho. 1 & $       \mathbf{0.019\pm   0.006}$ & $\mathbf{0.00422\pm8e-5}$ & $     0.665\pm   0.009$ & $    0.210\pm   0.002$\\
 & Ortho. 2 & $       0.046\pm   0.005$ & $0.0271\pm0.0017$ & $     0.688\pm    0.018$ & $    0.197\pm     0.010$\\
\hline
	segment & 1 vs. 1 & $        0.04\pm  5.9e-06$ & $0.0336\pm4e-4$ & $     \mathbf{0.911\pm   0.009}$ & $   \mathbf{0.0987\pm   0.0057}$\\
 & 1 vs. rest & $       0.012\pm   0.0042$ & $0.0094\pm2e-4$ & $     0.868\pm   0.010$ & $    0.144\pm   0.004$\\
 & ECC & $       0.016\pm   0.0052$ & $0.0124\pm4e-4$ & $     0.803\pm     0.040$ & $    0.179\pm     0.020$\\
	& Ortho. 1 & $       \mathbf{0.004\pm   0.005}$ & $\mathbf{0.00168\pm6e-5}$ & $     0.849\pm    0.015$ & $    0.166\pm    0.004$\\
 & Ortho. 2 & $        0.02\pm  2.9e-06$ & $0.0147\pm0.0012$ & $     0.880\pm    0.018$ & $    0.127\pm   0.008$\\
\hline
	shuttle & 1 vs. 1 & $       1.10\pm    0.03$ & $0.867 \pm 0.014$ & $     \mathbf{0.796\pm    0.013}$ & $   \mathbf{0.0824\pm   0.0017}$\\
 & 1 vs. rest & $       0.33\pm    0.01$ & $0.185\pm0.003$ & $     0.605\pm     0.010$ & $    0.1341\pm  0.0006$\\
 & ECC & $       0.42\pm    0.01$ & $0.265\pm0.011$ & $     0.535\pm     0.120$ & $    0.144\pm    0.026$\\
	& Ortho. 1 & $       \mathbf{0.183\pm   0.005}$ & $\mathbf{0.042\pm0.001}$ & $     0.593\pm   0.006$ & $    0.131\pm   0.002$\\
 & Ortho. 2 & $        0.48\pm    0.03$ & $0.31\pm0.03$ & $     0.710\pm    0.095$ & $    0.101\pm    0.024$\\
\hline
urban & 1 vs. 1 & $       0.031\pm   0.003$ & $0.0185\pm 1e-4$ & $     0.693\pm    0.026$ & $    \mathbf{0.188\pm   0.006}$\\
 & 1 vs. rest & $       \mathbf{0.007\pm   0.005}$ & $0.0052\pm 4e-4$ & $     0.667\pm    0.018$ & $    0.204\pm   0.004$\\
 & ECC & $       0.009\pm   0.003$ & $0.0068 \pm 4e-4$ & $     0.647\pm    0.031$ & $    0.210\pm   0.008$\\
 & ortho. 1 & $       \mathbf{0.007\pm   0.005}$ & $\mathbf{0.00064 \pm 4e-5}$ & $     0.674\pm    0.016$ & $    0.206\pm   0.004$\\
	& ortho. 2 & $       0.014\pm   0.005$ & $0.0082\pm 6e-4$ & $     \mathbf{0.693\pm    0.017}$ & $    0.198\pm   0.006$\\
\hline
	usps & 1 vs. 1 & $       0.63\pm    0.01$ & $0.347\pm0.005$ & $     \mathbf{0.898\pm     0.010}$ & $   \mathbf{0.0827\pm   0.0022}$\\
 & 1 vs. rest & $       0.152\pm   0.004$ & $0.0704\pm9e-4$ & $     0.840\pm   0.007$ & $    0.112\pm   0.003$\\
 & ECC & $       0.205\pm   0.005$ & $0.112\pm0.005$ & $     0.769\pm    0.021$ & $    0.1416\pm    0.006$\\
	& Ortho. 1 & $         \mathbf{0.1\pm  2.1e-05}$ & $\mathbf{0.0096\pm5e-4}$ & $     0.815\pm   0.009$ & $    0.132\pm   0.002$\\
 & Ortho. 2 & $       0.30\pm    0.02$ & $0.16\pm0.01$ & $      0.846\pm    0.015$ & $    0.112\pm   0.004$\\
\hline
	vehicle & 1 vs. 1 & $       0.002\pm   0.004$ & $0.00436\pm8e-5$ & $      \mathbf{0.685\pm    0.041}$ & $    \mathbf{0.245\pm    0.011}$\\
 & 1 vs. rest & $           0$ & $0.00142\pm6e-5$ & $     0.654\pm    0.037$ & $    0.263\pm   0.006$\\
 & ECC & $           0$ & $0.00143\pm8e-5$ & $     0.599\pm    0.049$ & $    0.279\pm    0.013$\\
	& Ortho. 1 & $           0$ & $\mathbf{0.00043\pm3e-5}$ & $     0.656\pm    0.038$ & $    0.263\pm   0.007$\\
 & Ortho. 2 & $           0$ & $0.0014\pm0.0001$ & $     0.636\pm    0.042$ & $    0.263\pm    0.019$\\
\hline
\end{tabular}

	}
\end{table*}

\begin{table*}
{\revision
\caption{Total classification time, solution time, uncertainty coefficient and Brier score for seven different datasets using five different coding matrices: 1 vs. 1, 1 vs. the rest, random, orthogonal with no zeros, and orthogonal with zeros. A support vector machine is used as the base binary classifier.}\label{class_results_svm}
	\begin{tabular}{ll|llll}
\hline
Dataset & Method & time (s) & sol. only (s) & U.C. & Brier score \\
\hline\hline
	pendigits & 1 vs. 1 & $       1.07\pm     0.14$ & $0.409\pm0.006$ & $     \mathbf{0.985\pm   0.003}$ & $   \mathbf{0.0319\pm   0.0024}$\\
	& 1 vs. rest & $       \mathbf{0.84\pm    0.10}$ & $0.082\pm0.002$ & $     0.981\pm   0.003$ & $   0.0361\pm   0.0034$\\
 & ECC & $       3.20\pm     0.86$ & $0.13\pm0.01$ & $     0.975\pm   0.004$ & $   0.0412\pm   0.0032$\\
	& ortho. 1 & $       2.13\pm     0.89$ & $\mathbf{0.013\pm0.002}$ & $     0.979\pm   0.004$ & $   0.0382\pm   0.0026$\\
	& ortho. 2 & $       1.17\pm     0.28$ & $0.20\pm0.01$ & $      0.982\pm   0.004$ & $   0.0354\pm   0.0034$\\
\hline
	sat & 1 vs. 1 & $       \mathbf{1.39\pm     0.35}$ & $0.077\pm0.009$ & $     \mathbf{0.800\pm     0.010}$ & $    \mathbf{0.145\pm   0.003}$\\
 & 1 vs. rest & $       1.70\pm     0.54$ & $0.028\pm0.005$ & $     0.786\pm   0.009$ & $    0.153\pm   0.003$\\
 & ECC & $       3.2\pm      1.6$ & $0.04\pm0.01$ & $     0.787\pm    0.011$ & $    0.152\pm   0.004$\\
	& ortho. 1 & $       3.8\pm        1.0$ & $\mathbf{0.008\pm0.003}$ & $     0.792\pm    0.011$ & $    0.149\pm   0.003$\\
 & ortho. 2 & $        1.79\pm     0.52$ & $0.034\pm0.007$ & $     0.789\pm   0.009$ & $    0.150\pm   0.004$\\
\hline
	segment & 1 vs. 1 & $       0.18\pm    0.05$ & $0.034\pm0.001$ & $     0.923\pm   0.007$ & $   \mathbf{0.0882\pm   0.0053}$\\
	& 1 vs. rest & $       \mathbf{0.11\pm    0.03}$ & $0.0102\pm0.0005$ & $      0.919\pm   0.007$ & $   0.0938\pm   0.0051$\\
 & ECC & $       0.13\pm    0.07$ & $0.014\pm0.001$ & $     0.915\pm    0.013$ & $   0.0938\pm   0.0071$\\
	& ortho. 1 & $       0.16\pm    0.07$ & $\mathbf{0.0018\pm0.0001}$ & $     \mathbf{0.925\pm   0.008}$ & $   0.0890\pm   0.0048$\\
 & ortho. 2 & $       0.11\pm    0.03$ & $0.015\pm0.001$ & $     0.919\pm    0.012$ & $   0.0883\pm    0.0050$\\
\hline
	shuttle & 1 vs. 1 & $       6.3\pm        1.0$ & $0.98\pm0.06$ & $      \mathbf{0.982\pm   0.003}$ & $   \mathbf{0.0182\pm   0.0015}$\\
	& 1 vs. rest & $       \mathbf{6.0\pm      1.6}$ & $0.26\pm0.03$ & $     0.978\pm   0.006$ & $   0.0215\pm   0.001$\\
 & ECC & $       12.4\pm      5.7$ & $0.43\pm0.10$ & $     0.878\pm     0.210$ & $   0.0731\pm      0.100$\\
	& ortho. 1 & $      10.0\pm      4.7$ & $\mathbf{0.09\pm0.03}$ & $     0.974\pm    0.003$ & $   0.0222\pm    0.0010$\\
 & ortho. 2 & $       6.6\pm      1.6$ & $0.40\pm0.04$ & $     0.978\pm   0.002$ & $   0.0230\pm   0.0068$\\
\hline
	urban & 1 vs. 1 & $       0.41\pm     0.21$ & $0.222\pm 0.003$ & $     \mathbf{0.726\pm    0.035}$ & $\mathbf{0.170\pm   0.009}$\\
	& 1 vs. rest & $        0.26\pm    0.10$ & $0.0059\pm7e-4$ & $     0.708\pm    0.038$ & $    0.176\pm    0.011$\\
	& ECC & $       0.71\pm     0.31$ & $0.0085\pm 0.0011$ & $     0.711\pm     0.030$ & $    0.178\pm   0.009$\\
	& ortho. 1 & $       0.79\pm     0.24$ & $\mathbf{0.0014\pm3e-4}$ & $     0.723\pm    0.023$ & $     0.173\pm   0.009$\\
	& ortho. 2 & $       \mathbf{0.22\pm     0.15}$ & $0.0088\pm0.0011$ & $     0.715\pm    0.026$ & $    0.172\pm   0.009$\\
\hline
	usps & 1 vs. 1 & $        33.9\pm       17.0$ & $0.42\pm0.02$ & $     \mathbf{0.929\pm   0.006}$ & $   \mathbf{0.0664\pm   0.0023}$\\
	& 1 vs. rest & $      \mathbf{22.9\pm      7.6}$ & $0.110\pm0.009$ & $     0.921\pm   0.005$ & $   0.0732\pm    0.0020$\\
 & ECC & $      73.0\pm       29.0$ & $0.150\pm0.009$ & $     0.915\pm    0.006$ & $    0.0754\pm   0.0022$\\
	& ortho. 1 & $      70.1\pm       29.0$ & $\mathbf{0.018\pm0.003}$ & $     0.922\pm   0.006$ & $   0.0712\pm   0.0018$\\
 & ortho. 2 & $      34.8\pm       16.0$ & $0.21\pm0.02$ & $     0.920\pm   0.008$ & $   0.0707\pm   0.0027$\\
\hline
	vehicle & 1 vs. 1 & $       0.047\pm    0.013$ & $0.00465\pm8e-5$ & $     0.635\pm    0.023$ & $    \mathbf{0.272\pm   0.007}$\\
 & 1 vs. rest & $       0.055\pm    0.016$ & $0.0016\pm0.001$ & $     0.625\pm    0.033$ & $    0.277\pm   0.009$\\
 & ECC & $       0.053\pm    0.024$ & $0.0017\pm0.0002$ & $     0.610\pm    0.061$ & $    0.282\pm    0.011$\\
	& ortho. 1 & $        0.050\pm    0.018$ & $\mathbf{0.00050\pm3e-5}$ & $     0.621\pm    0.032$ & $    0.277\pm    0.009$\\
	& ortho. 2 & $       \mathbf{0.042\pm   0.006}$ & $0.00155\pm9e-5$ & $     \mathbf{0.639\pm    0.025}$ & $    0.278\pm   0.009$\\
\hline
\end{tabular}
}
\end{table*}

\begin{table*}
{\revision
\caption{Solution time, uncertainty coefficient and Brier score for seven different datasets using five different coding matrices: 1 vs. 1, 1 vs. the rest, random, orthogonal with no zeros, and orthogonal with zeros. A piecewise linear classifier is used as the base binary classifier.}\label{class_results_acc}
	\begin{tabular}{ll|llll}
\hline
Dataset & Method & time (s) & sol. only (s) & U.C. & Brier score \\
\hline\hline
	pendigits & 1 vs. 1 & $       1.71\pm    0.08$ & $0.45\pm0.02$ & $     \mathbf{0.977\pm   0.005}$ & $   \mathbf{0.0383\pm   0.003}$\\
	& 1 vs. rest & $       \mathbf{0.62\pm    0.02}$ & $0.088\pm0.004$ & $     0.967\pm   0.006$ & $   0.0539\pm   0.0021$\\
 & ECC & $       0.77\pm    0.02$ & $0.14\pm0.01$ & $     0.955\pm    0.011$ & $   0.0603\pm   0.0061$\\
	& ortho. 1 & $       0.64\pm    0.01$ & $\mathbf{0.0122\pm0.0005}$ & $     0.961\pm   0.006$ & $   0.0560\pm   0.0037$\\
 & ortho. 2 & $       1.3\pm      0.1$ & $0.21\pm0.02$ & $     0.969\pm   0.007$ & $   0.0471\pm   0.0033$\\
	\hline
	sat & 1 vs. 1 & $       1.97\pm    0.07$ & $0.098\pm0.02$ & $     \mathbf{0.783\pm   0.009}$ & $    \mathbf{0.159\pm   0.005}$\\
	& 1 vs. rest & $       \mathbf{1.17\pm    0.03}$ & $0.035\pm0.007$ & $     0.768\pm    0.012$ & $    0.168\pm   0.003$\\
 & ECC & $       1.54\pm    0.05$ & $0.045\pm0.01$ & $     0.765\pm    0.013$ & $    0.165\pm    0.004$\\
	& ortho. 1 & $       1.50\pm    0.04$ & $\mathbf{0.010\pm0.004}$ & $     0.776\pm   0.009$ & $    0.162\pm   0.004$\\
 & ortho. 2 & $       1.6\pm     0.2$ & $0.047\pm0.01$ & $     0.763\pm   0.009$ & $    0.169\pm   0.010$\\
	\hline
	segment & 1 vs. 1 & $        0.170\pm   0.005$ & $0.0353\pm4e-4$ & $     \mathbf{0.911\pm    0.011}$ & $   \mathbf{0.096\pm   0.005}$\\
	& 1 vs. rest & $       \mathbf{0.099\pm   0.0032}$ & $0.0104\pm4e-4$ & $     0.883\pm    0.019$ & $    0.119\pm   0.004$\\
 & ECC & $       0.113\pm   0.005$ & $0.015\pm0.001$ & $     0.888\pm    0.026$ & $     0.116\pm   0.010$\\
	& ortho. 1 & $       \mathbf{0.099\pm   0.003}$ & $\mathbf{0.00190\pm5e-5}$ & $     0.896\pm    0.011$ & $    0.115\pm   0.005$\\
 & ortho. 2 & $        0.15\pm    0.01$ & $0.0160\pm7e-4$ & $     0.910\pm    0.011$ & $    0.103\pm   0.007$\\
	\hline
	shuttle & 1 vs. 1 & $       4.398\pm    0.093$ & $0.90\pm0.03$ & $     \mathbf{0.981\pm   0.010}$ & $   0.0274\pm    0.0110$\\
	& 1 vs. rest & $       \mathbf{2.51\pm    0.04}$ & $0.217\pm0.006$ & $     0.967\pm    0.028$ & $   0.0315\pm   0.0083$\\
 & ECC & $       2.89\pm     0.06$ & $0.28\pm0.02$ & $     0.972\pm    0.005$ & $   0.0313\pm   0.0044$\\
	& ortho. 1 & $       2.63\pm    0.04$ & $\mathbf{0.045\pm0.001}$ & $     0.976\pm   0.002$ & $    \mathbf{0.0261\pm   0.0010}$\\
 & ortho. 2 & $       3.7\pm     0.3$ & $0.35\pm0.03$ & $     0.976\pm   0.004$ & $   0.0270\pm   0.0043$\\
	\hline
	urban & 1 vs. 1 & $        0.94\pm    0.02$ & $0.023\pm0.001$ & $     \mathbf{0.724\pm    0.019}$ & $    \mathbf{0.172\pm    0.009}$\\
	& 1 vs. rest & $       \mathbf{0.23\pm    0.01}$ & $0.005\pm0.001$ & $     0.698\pm    0.032$ & $    0.184\pm    0.011$\\
 & ECC & $       0.314\pm   0.008$ & $0.008\pm0.001$ & $     0.692\pm    0.028$ & $    0.184\pm   0.006$\\
	& ortho. 1 & $       0.31\pm    0.01$ & $\mathbf{0.0012\pm4e-4}$ & $     0.717\pm    0.022$ & $    0.176\pm   0.008$\\
 & ortho. 2 & $       0.44\pm    0.03$ & $0.011\pm0.001$ & $      0.719\pm    0.034$ & $    0.176\pm    0.015$\\
	\hline
	usps & 1 vs. 1 & $      14.4\pm     0.2$ & $0.41\pm0.02$ & $     \mathbf{0.914\pm   0.005}$ & $   \mathbf{0.075\pm   0.002}$\\
	& 1 vs. rest & $       \mathbf{6.2\pm     0.1}$ & $0.08\pm0.01$ & $     0.897\pm   0.007$ & $     0.101\pm   0.002$\\
 & ECC & $       7.5\pm     0.1$ & $0.14\pm0.02$ & $     0.881\pm   0.006$ & $   0.095\pm   0.003$\\
	& ortho. 1 & $       7.3\pm      0.1$ & $\mathbf{0.014\pm0.004}$ & $     0.897\pm   0.006$ & $   0.089\pm   0.002$\\
 & ortho. 2 & $      12\pm        1$ & $0.20\pm0.02$ & $     0.899\pm   0.008$ & $   0.084\pm   0.003$\\
	\hline
	vehicle & 1 vs. 1 & $       0.017\pm   0.005$ & $0.0044\pm1e-4$ & $     \mathbf{0.628\pm    0.038}$ & $      \mathbf{0.273\pm    0.007}$\\
 & 1 vs. rest & $       0.017\pm   0.005$ & $0.00156\pm8e-5$ & $      0.607\pm    0.036$ & $     0.282\pm   0.007$\\
 & ECC & $        0.02\pm  2.9e-06$ & $0.00158\pm5e-5$ & $     0.602\pm    0.067$ & $    0.283\pm    0.014$\\
	& ortho. 1 & $       \mathbf{0.015\pm   0.005}$ & $\mathbf{0.00046\pm1e-5}$ & $     0.614\pm    0.026$ & $    0.281\pm   0.007$\\
 & ortho. 2 & $       0.016\pm   0.005$ & $0.0015\pm1e-4$ & $      0.597\pm    0.041$ & $    0.287\pm    0.011$\\
\hline
\end{tabular}

}
\end{table*}

Orthogonal error-correcting codes were tested on 
{\revision seven} different datasets:
two for digit recognition--``pendigits'' \citep{Alimoglu1996} and
``usps'' \citep{Hull1994}; the space shuttle control dataset--``shuttle''
\citep{King_etal1995}; 
{\revision an urban land classification dataset--``urban'' \citep{Johnson2013};}
a similar one for satellite land classification--``sat''; 
a dataset for patterned image recognition--``segment'';
and a dataset for vehicle recognition--``vehicle'' \citep{Siebert1987}.
The last three are borrowed from the ``statlog'' project \citep{King_etal1995,Michie_etal1994}.

{\revision
Two types of orthogonal ECCs were tested: the first type described in
Section \ref{construction}, with no zeros in the codes, and the second type which includes
zeros.
These were compared with three other methods: one-versus-one, one-versus-the-rest,
and random ECCs with the same length of coding vector (number of columns), $m$,
as the orthogonal matrices of the first type.
The 1 vs. rest multi-class as well as the random ECCs were solved using 
the same type of constrained linear least squares method as used for the
second type of orthogonal ECC \citep{Lawson_Hanson1995}.
By enforcing the normality constraints using a Lagrange multiplier,
1 vs. 1 may be solved with a simple (unconstrained) linear equation solver \citep{Wu_etal2004}.}

{\revision
Three types of binary classifier were used: logistic regression \citep{Michie_etal1994},
support vector machines \citep{Mueller_etal2001},
and a peicewise-linear classifer \citep{Mills2018}.
Logistic regression classifiers were trained using LIBLINEAR \citep{Fan_etal2008}.}

{\revision Support vector machines (SVMs) were trained using
LIBSVM \citep{Chang_Lin2011}.}
Partitions were trained separately then combined by finding the union of
sets of support vectors for each partition.
By indexing into the combined list of support vectors, the algorithms are
optimized in both space and time \citep{Chang_Lin2011}.
For SVM, the same parameters were used for all multi-class methods and
for all partitions (matrix columns).
All datasets were trained using  ``radial basis function'' (Gaussian)
kernels of differing widths.

{\revision
LIBSVM was also used to train an intermediate model from which an often faster 
piecewise-linear classifier \citep{Mills2018} was trained.
It was thought that this classifier would provide a better use-case for orthogonal
ECCs than either of the other two.
The single parameter for this algorithm--the number of border vectors--was set 
the same for each dataset as used in \citet{Mills2018} for the 1 vs. 1. 
For the other multi-class algorithms, the number of border vectors
was doubled for small values (under 100) and increased by fifty percent
for larger values to account for the more complex decision function
created by using more classes in each binary classifier.
Multi-class classifiers were designed, trained and applied using the
framework provided within libAGF \citep{Mills2018,Mills2011,Mills2018a}
}

Results are shown in {\revision Tables \ref{class_results_lin}, \ref{class_results_svm},
and \ref{class_results_acc}.}
Confidence limits represent standard deviations over 10 trials using
different, randomly chosen coding matrices.
For each trial, datasets were randomly separated into 70\% training and 30\%
test.
``U.C'' stands for uncertainty
coefficient, a skill score based on Shannon's channel capacity
that has many advantage over simple
fraction of correct guesses or ``accuracy''
\citep{Mills2011,Shannon,Press_etal1992}.
Probabilities are validated with the Brier score 
which is root-mean-square
error measured against the truth of the class as a 0 or 1 value
\citep{Brier1950,Jolliffe_Stephenson2003}.

For all of the datasets tested, orthogonal ECCs provide a small but
significant improvement over random ECCs in both classification
accuracy and in the accuracy of the conditional probabilities.
This is in line with the literature as in \citet{Dietterich_Bakiri1995,Windeatt_Ghaderi2002}.
{\revision
Improvements range from 0.4\% to 17.5\% relative (0.004 to 0.139 absolute) in 
uncertainty coefficient
and 0.7\% to 10.7\% in Brier score.}
Results are also more consistent for the orthogonal ECCs as given by the
calculated error bars.

{\revision
Also as expected, solution times are
extremely fast for the first type of orthogonal ECC.
In many cases the times are an order-of-magnitude better than the next
fastest method.
Depending on the problem and classification method, this may or may not
be significant.
Since SVM is a relatively slow classifier, solution times
are a minor portion of the total.
For the logistic regression classifier,
solving the constrained optimization problem for the probabilities
typically comprises the bulk of classification times.
Oddly, the solver for the 1 vs. 1 method is the slowest by a wide margin,
even though it's a simple (unconstrained) linear solver \citep{Wu_etal2004}.
This could potentially be improved by using a faster solver \citep{Press_etal1992} or by employing
the iterative method given in \citet{Wu_etal2004}.

The two types of orthogonal ECCs were quite close in accuracy,
with sometimes one taking the lead and sometimes the other.
For the linear classifier, the second type was always more accurate while the first
type was faster.
Since it admits zeros, the decision boundaries are usually simpler--see below.
For both the SVM and the piecewise linear classifier, skill scores were very similar,
differing by at most 2.9\% relative, 0.018 absolute, in U.C. and 17\% in Brier score.
For the SVM, the second type was faster while for the piecewise linear classifier,
the first type was faster.
The explanation for this follows.

Unfortunately, there is one method that is consistently more accurate than the
orthogonal ECCs and this is 1 vs. 1.
The orthogonal ECCs only beat 1 vs. 1 three times out of 21 for the
uncertainty coefficient and one time out of 21 for the Brier score.
Improvements in uncertainty coefficient range from insignificant
to 0.6\% relative or 0.004 absolute.
The Brier score improved by 2.6\%.
Losses using linear classifiers were the worst, peaking at 14.6\% relative,
0.203 absolute, in uncertainty coefficient and 50\% in Brier score.
The results for logistic regression provide a vivid demonstration as to
why 1 vs. 1 works so well: because it partitions the classes into
``least-divisible units'', there are fewer training samples provided to
each binary classifier, the decision boundary is simpler and
a simpler classifier will work better

Nonetheless, there is a potential use case for our method.
Although orthogonal ECCs are less accurate than 1 vs. 1, they
don't lose much.
If they are also faster, then a speed improvement may be worth a small
hit in accuracy for some applications \citep{Mills2018}.
While 1 vs. 1 beats orthogonal ECCs by a healthy margin using linear
classifiers, the biggest loss in U.C. for SVM is only 1.5\% relative, 
0.011 absolute.
Losses for Brier score are somewhat worse,
peaking at 6.5\%.
Unfortunately, because the speed of a multi-class SVM is proportional
mainly to the total number of support vectors \citep{Mills2018},
orthogonal ECCs rarely provide much of a speed advantage.
What is needed is a constant-time--ideally very fast--non-linear classifier.
This is where the piecewise-linear classifier comes in.

For uncertainty coefficient, 1 vs. 1 was always better than orthogonal ECCs
when using the piecewise-linear classifier. 
Losses peak at 1.9 \% relative, 0.017 absolute.
For the Brier score, only one of the seven datasets showed an improvement
over 1 vs. 1 at 4.9 \%.
The worst loss was 39 \%.
Improvements in speed range from 1.1 \% to over 100 \%.
Much of the speed difference is simply the result of using fewer binary
classifiers.

The purpose of the piecewise linear classifier is to improve the speed of the
SVM.
This speed increase is better with orthogonal ECCs than with 1 vs. 1.
Orthogonal ECCs applied to piecewise linear classifiers 
are faster than the the fastest
SVM for five out of the seven datasets.
Speed often trades off from accuracy.
\citet{Mills2018} provides a procedure for determining whether it's worth
switching algorithms or not.
A similar analysis will not be repeated here due to time and space 
considerations, however whether any improvement in speed is worth
the consequent hit in accuracy will depend on the application.}

\section{Conclusions}

As predicted by recent literature, solving for multi-class using orthogonal ECCs was more accurate than the equivalent problem using random ECCs.
{\revision
Unfortunately, they were still unable to beat one-versus-one as an
effective multi-class method.
The author's own work suggests that the 1 vs. 1 classification almost always
works well regardless of the dataset \citep{Mills2018a}.
\citet{Hsu_Lin2002} find that 1 vs. 1 outperform both 1 vs. rest and random ECCs on a test
of ten different datasets using SVM.
One-versus-one is also used, often exclusively, with many statistical classification software packages.

There may still be room for further work, however, with the most likely fruitful
line of inquiry being, first,
on adaptive methods that use the data to figure out how best to go
from binary to multi-class.
In \citet{Mills2018a}, for instance, even though 1 vs. 1 was almost always
most accurate, there was one dataset that benefitted from a more
customized treatment.
Recent work has focused on both empirically-designed decision trees 
\citep{Cheong_etal2004,Lee_Oh2003,Benabdeslem_Bennani2006}
as well as empirically-designed ECCs 
\citep{Crammer_Singer2002,Zhou_etal2008,Zhong_Cheriet2013}.
Decision trees are the easiest to tackle because there are fewer possibilities
and because a tree can be built from either the top down or the bottm up.

A second potential area for future work is in multi-class methods integrated
with the base binary classifier, for instance with all the binary classifiers
being trained simultaneously \citep{Hsu_Lin2002}.
It stands to reason that more integrated multi-class methods would tend to
be more accurate than those, such the ones disussed in this note, that treat the
binary classifier as a ``black box'', since there can now be sharing of
information.

There is also a potential use case for orthogonal ECCs. 
If they are paired with a fast, non-linear binary classifier
with better than $O(N)$ performance, where $N$ is the number of training
samples,
orthogonal ECCs should almost always be faster than 1 vs. 1
while giving up little in accuracy.}
The algorithm presented here that solves for the probabilities is simple and elegant and
may suggest new directions in the search for more efficient and
accurate multi-class classification algorithms.
{\revision
Since it is fast it could help provide speed
improvements for such applications as real-time computer vision,
image processing, and voice-recognition.}

\appendix

\section*{acknowledgements}
Thanks to Chih-Chung Chan and Chih-Jen Lin of the National Taiwan University
for data from the LIBSVM archive and also to David Aha and the curators of
the UCI Machine Learning Repository for statistical classification datasets.

	The LIBSVM software libraries can be found: \url{https://www.csie.ntu.edu.tw/~cjlin/libsvm/}.
	The LIBLINEAR software libraries can be found: \url{https://www.csie.ntu.edu.tw/~cjlin/liblinear/}.
	Software for performing multi-class classification using orthogonal error correcting codes, and many others, can be found: \url{https://www.github.com/peteysoft/libmsci}.

\bibliographystyle{apa}


\bibliography{../../agf_bib,../orthogonal,../../pwl,../../svm_accel/svm_accel,../../datasets}   

\end{document}